\documentclass{article}

\pdfoutput=1
\usepackage[numbers]{natbib}

\usepackage[preprint]{_NeurIPS/neurips_2022}

\usepackage[toc,page]{appendix}

\usepackage{amsmath}
\usepackage{amssymb}

\usepackage{xcolor}

\usepackage{hyperref}

\usepackage{graphicx}
\usepackage{subcaption}
\usepackage{wrapfig}

\usepackage{booktabs}
\usepackage{rotating}

\usepackage{tcolorbox}

\usepackage[utf8]{inputenc} %
\usepackage[T1]{fontenc}    %
\usepackage{hyperref}       %
\usepackage{url}            %
\usepackage{booktabs}       %
\usepackage{amsfonts}       %
\usepackage{nicefrac}       %
\usepackage{microtype}      %
\usepackage{xcolor}         %

\usepackage{todonotes}
\usepackage{paralist}
\usepackage[font=small, labelfont=bf]{caption}

\usepackage[export]{adjustbox}

\hypersetup{
    colorlinks = true,
    citecolor=blue,
    linkcolor=blue,
    urlcolor=blue
}

\definecolor{darkgreen}{rgb}{0.0, 0.5, 0.0}
\definecolor{blue}{rgb}{0.0, 0.47, 0.75}
\definecolor{dartmouthgreen}{rgb}{0.05, 0.5, 0.06}
\definecolor{drab}{rgb}{0.59, 0.44, 0.09}
\definecolor{navyblue}{rgb}{0.0, 0.0, 0.5}

\newcommand{\KL}{\mathbb{KL}}

\newcommand{\vc}{\mathbf{c}}

\newcommand{\vu}{\mathbf{u}}

\newcommand{\vx}{\mathbf{x}}

\newcommand{\vz}{\mathbf{z}}

\newcommand{\vI}{\mathbf{I}}

\newcommand{\vX}{\mathbf{X}}

\newcommand{\vtheta}{\mathbf{\theta}}

\newcommand{\vphi}{\mathbf{\phi}}

\newcommand{\cN}{\mathcal{N}}

\newcommand{\bE}{\mathbb{E}}

\newcommand{\bR}{\mathbb{R}}

\newcommand{\vzero}{\mathbf{0}}

\title{Few-Shot Diffusion Models}

\author{%
  Giorgio Giannone \\
  Technical University of Denmark \\
  \texttt{gigi@dtu.dk} \\
  \And
  Didrik Nielsen \\
  Norwegian Computing Center\\
  \texttt{didrik@nr.no} \\
  \And
  Ole Winther \\
  Technical University of Denmark \\ University of Copenhagen \\
  \texttt{olwi@dtu.dk} \\
}

\begin{document}

\maketitle
\begin{abstract}
    Denoising diffusion probabilistic models (DDPM) are powerful hierarchical latent variable models with remarkable sample generation quality and training stability.
    These properties can be attributed to parameter sharing in the generative hierarchy, as well as a parameter-free diffusion-based inference procedure.
    In this paper, we present Few-Shot Diffusion Models (FSDM), a framework for few-shot generation leveraging conditional DDPMs. FSDMs are trained to adapt the generative process conditioned on a small set of images from a given class by aggregating image patch information using a set-based Vision Transformer (ViT). At test time, the model is able to generate samples from previously unseen classes conditioned on as few as 5 samples from that class.
    We empirically show that FSDM can perform few-shot generation and transfer to new datasets. %
    We benchmark variants of our method on complex vision datasets for few-shot learning and compare to unconditional and conditional DDPM baselines.
    Additionally, we show how conditioning the model on patch-based input set information improves training convergence.
    
\end{abstract}

\section{Introduction}

\begin{figure}[ht]
    \centering
    \includegraphics[trim={0, 82pt, 0, 0}, clip, width=\linewidth]{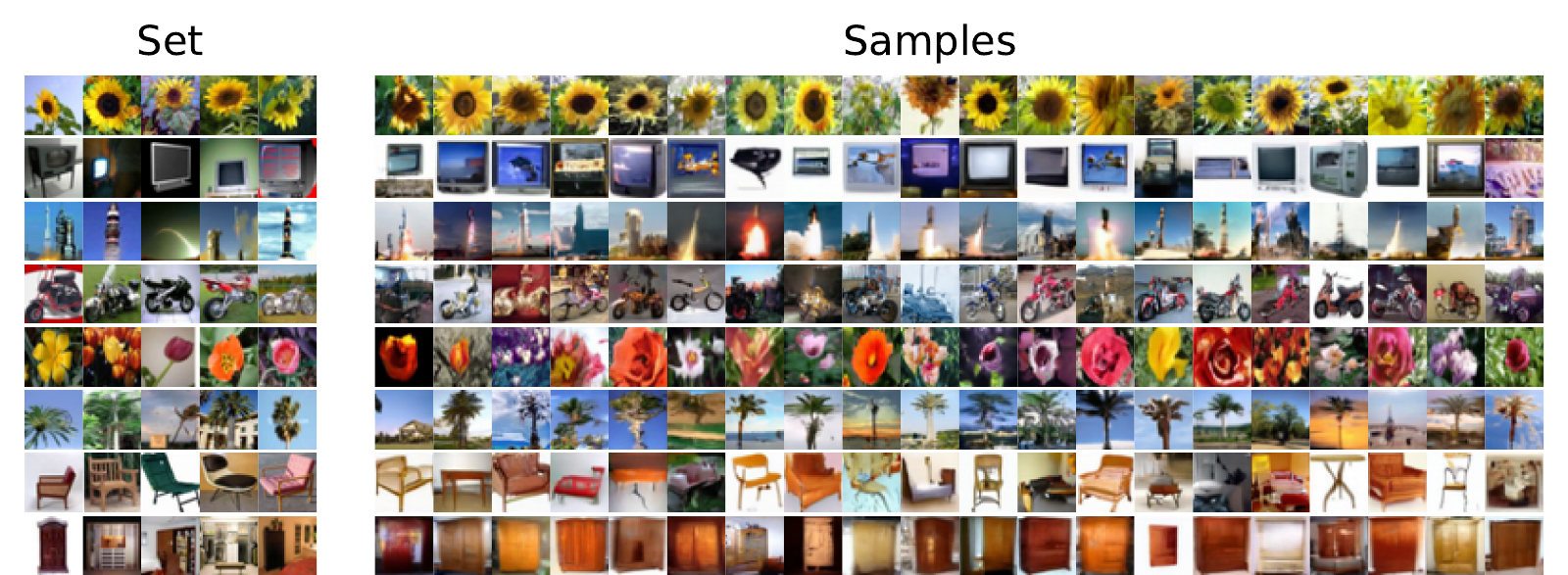}
    \caption{Set (left) and conditional samples (right) on CIFAR100 using a Few-Shot Diffusion Models.
    FSDM can extract content information from a handful of realistic examples and generate rich and complex samples from a variety of conditional distributions. More samples in Appendix Fig.~\ref{fig:intro_samples_cifar100_app}.}
    \label{fig:intro_samples_cifar100}
\end{figure}

Humans are exceptional few-shot learners able to grasp concepts and function of objects never encountered before~\citep{lake2011one, tenenbaum2011grow, lake2015human}. 
This is because we build internal models of the world so we can combine our prior knowledge about object appearance and function to make well-educated inferences from very little data~\citep{tenenbaum1999bayesian,lake2017building,ullman2020bayesian}. 
In contrast, traditional machine learning systems have to be trained tabula rasa and therefore need orders of magnitude more data. 

A particularly challenging problem is \emph{few-shot adaptation in generative latent variable models}~\citep{edwards2016towards,reed2017few, rezende2016one, bartunov2018few}.
Few-shot generation has been limited to simple datasets and shallow tasks, using handcrafted aggregation and conditioning mechanisms.

Generative models~\citep{kingma2013auto, rezende2014stochastic, kingma2019introduction, dinh2016density, rezende2015variational, papamakarios2021normalizing, goodfellow2014generative, arjovsky2017wasserstein, mohamed2016learning, oord2016pixel, sohl2015deep, song2019} for high-dimensional, complex data modalities like images have been a challenge for the machine learning community.
Large vision~\citep{vahdat2020NVAE, child2020very, maaloe2019biva, chen2017pixelsnail, child2019generating, brock2018large}, and language~\citep{devlin2018bert, brown2020language} models have greatly increased our capacity to process unstructured data, opening the door for multimodal and semantic generation~\citep{ramesh2022hierarchical, nichol2021glide, blattmann2022retrieval, ramesh2021zero}.

Recently diffusion models~\citep{song2019, ho2020} have shown impressive generative performance for vision~\citep{nichol2021, ho2022video, ho2022cascaded}, language~\citep{hoogeboom2021argmax, austin2021structured}, speech~\citep{kong2020diffwave}, biological data~\citep{hoogeboom2022equivariant, luo2021diffusion, xu2022geodiff}, and multimodal~\citep{nichol2021glide, ramesh2022hierarchical} generation, providing an important step toward general and stable pure generative models.
Unconditional diffusion models are expressive likelihood-based density estimators with high sample quality~\citep{kingma2021density}. 
This expressivity arises from the Monte Carlo (layer) sampling that we can perform during training thanks to the special structure of the forward process: a parameter-free diffusion process for which the posterior can be computed at each step in closed form~\citep{ho2020}.

\begin{wrapfigure}{r}{0.3\textwidth}
\vspace*{-\baselineskip}
\centering
    \includegraphics[width=\linewidth]{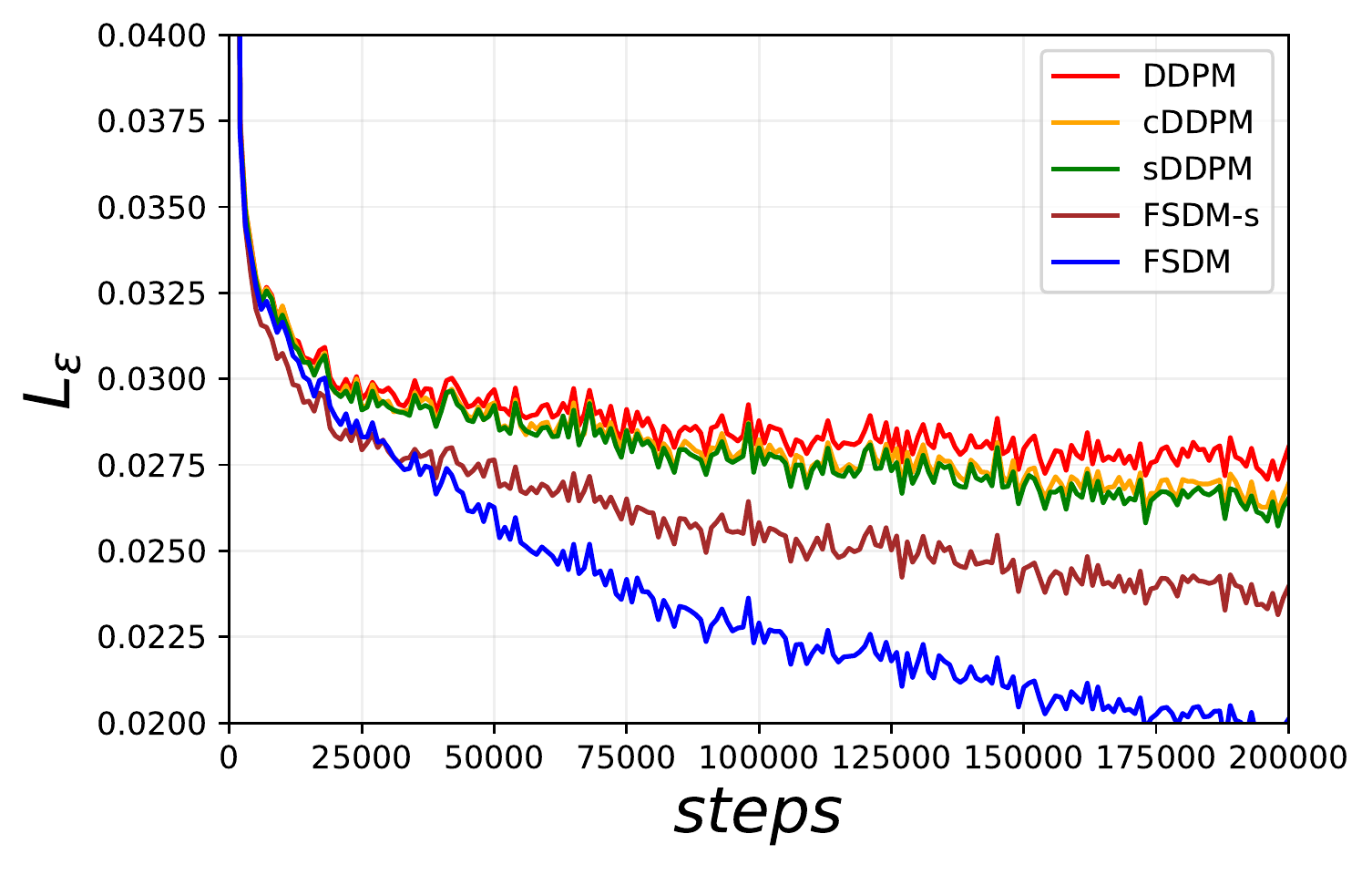}
    \caption{Estimated $L_{\epsilon}$ per layer on CIFAR100 during training.
    FSDM is data efficient during training and can denoise the data better and faster than unconditional and conditional DDPM baselines.}
    \label{fig:FSDM_mse_cifar100mix}
\vspace*{-\baselineskip}
\end{wrapfigure}
However such effectiveness is at the cost of posterior flexibility and absence of latent space structure.
For this reason the few-shot capacities of this class of models is largely unexplored and conditional adaptation is challenging.

Conditioning mechanisms have been proposed for test-time adaptation~\citep{choi2021ilvr}, low-shot attribute generation~\citep{sinha2021d2c}, and class information~\citep{nichol2021improved, song2020score}, but not to generate complex, new classes or objects never encounter during training. 
When dealing with complex novel objects, simple conditioning mechanisms fail, demonstrating a need for more expressive approaches for few-shot generation.

In this work we aim to study adaptation mechanisms and improve few-shot generation in latent variable models on realistic and complex visual data (Fig.~\ref{fig:intro_samples_cifar100}). 
The setting we consider is that of learning from a large quantity of homogeneous sets, where each set is an un-ordered collections of samples of one concept or class. 
At test time, the model will be provided with sets of concepts never encountered during training.
We consider explicit conditioning in a hierarchical formulation, where \emph{global variables carry information about the set at hand}.
The conditional hierarchical model can naturally represent a family of generative models, each specified by a different conditioning set-level variable. 
We formulate FSDM using vision transformers~\citep{dosovitskiy2020image} and diffusion models~\citep{ho2020, nichol2021}. We propose to process the input set in patches and condition the generative model with a learnable attention mechanism using a tokenized representation for the input set. 

\textbf{Our contributions} are
\begin{compactitem}
    \item a new framework to perform few-shot generation for realistic sets of images in the DDPM framework. 
    \item Learnable Attentive Conditioning (LAC), a conditioning mechanism where the input set is processed as a collection of patches and used to condition a DDPM through attention between sample-level and set-level variables.
    \item Experimental evidence that our model speeds up training, increases sample quality and variety, and improves transfer for conditional and few-shot generation compared to relevant unconditional and conditional DDPM-based baselines.
\end{compactitem}
\section{Background}

\paragraph{Few-Shot Generation.}
Few-shot generation is the task of adapting quickly to new classes or objects at test time given a small amount of instances from a novel category. In standard few-shot learning~\citep{lake2011one,lake2015human, hospedales2020meta}, given a support set $\vX_s = \{\vx_s\}^S_{s=1}$ and a query sample $\vx_q$, 
we condition a learner on $\vX_s$ and predict on $\vx_q$ with a model of the form $p(\vx_q | f_{\phi}(\vX_s))$. The conditioning can be explicit on the representations~\citep{garnelo2018neural, garnelo2018conditional, vinyals2016matching, snell2017prototypical, oreshkin2018tadam, sung2018learning, shyam2017attentive} or implicit on the parameters like in meta-learning~\citep{schaul2010metalearning, hochreiter2001learning, finn2017model, grant2018recasting, ravi2018amortized} and optimization~\citep{rusu2018meta, ravi2016optimization, andrychowicz2016learning}.
Few-shot generation~\citep{lake2015human} borrows a similar setting but for the more challenging task to generate objects given few samples of that object at test time. In particular, given context information $\vX$, we want to learn a conditional generative model that adapts quickly to new objects. There are two main ways to do this: learn a set-based generative model $p(\vX) = \int p(\vX | \vc) p(\vc) d\vc$ and perform few-shot generation as a downstream task leveraging the per-set posterior $q_{\phi}(\vc | \vX_{\texttt{new}})$ as in~\citep{edwards2016towards, hewitt2018variational, giannone2021hierarchical}.
Alternatively, we can learn a conditional model of the form $p(\vx | \vX) = \int p(\vx | \vc) p(\vc | \vX) d\vc$ similarly to~\citep{bartunov2018few, rezende2016one, reed2017few} and perform few-shot generation using the model directly for $p(\vc | \vX_{\texttt{new}})$. We consider the deterministic case of this, where the encoder $p(\vc | \vX) = \delta(\vc - h_{\phi}(\vX))$ is a deterministic set-based neural network $\vc \leftarrow h_{\phi}(\vX)$.

\paragraph{Diffusion Denoising Probabilistic Models (DDPM).}
Let $\vx_0$ denote the observed data which is either continuous $\vx_0 \in \bR^D$ or discrete $\vx_0 \in \{0,...,255\}^D$. Let $\vx_1, ..., \vx_T$ denote $T$ latent variables in $\bR^D$.
We now introduce, the \emph{forward or diffusion process} $q$, the \emph{reverse or generative process} $p_{\theta}$ and the objective $L$.
The forward or diffusion process $q$ is defined as~\citep{ho2020}:
\begin{align}
    q(\vx_{1:T} | \vx_0) = \prod_{t=1}^T q(\vx_t | \vx_{t-1}), \quad q(\vx_t | \vx_{t-1}) = \cN(\vx_t | \sqrt{1-\beta_t}~\vx_{t-1}, \beta_t I)\label{eq:ddpm_inference_main}
\end{align}
The beta schedule $\beta_1, \beta_2, ..., \beta_T$ is chosen such that the final latent image $\vx_T$ is nearly Gaussian noise.
The generative or inverse process $p_{\theta}$ is defined as:
\begin{align}
    p_{\theta}(\vx_{0:T}) = p(\vx_T)\prod_{t=1}^T p_{\theta}(\vx_{t-1}|\vx_t),
     \quad p_{\theta}(\vx_{t-1}|\vx_t) = \cN(\vx_{t-1}|\mu_{\theta}(\vx_t, t), \sigma_t^2 I),
\end{align}
where $p(\vx_T)= \cN(\vx_T| 0, I)$, and $\sigma^2_t$ often is fixed (e.g. to $\sigma^2_t = \beta_t$).
The neural network $\mu_{\theta}(\vx_t, t)$ is shared among all time steps and is conditioned on $t$. 
The model is trained with a re-weighted version of the ELBO that relates to denoising score matching \citep{song2019}.
The negative ELBO $L$ can be written as
\begin{align}
    L
    = \bE_q\left[- \log \dfrac{p_{\theta}(\vx_{0:T})}{q(\vx_{1:T} | \vx_0)} \right]
    = L_0 + \sum_{t=2}^{T} L_{t-1} + L_T,
    \label{eq:loss_ddpm_main}
\end{align}

where $L_0 = \bE_{q(\vx_1|\vx_0)} \left[- \log p(\vx_0 |\vx_1) \right]$ is the likelihood term (parameterized by a discretized Gaussian distribution)
and, if $\beta_1,...\beta_T$ are fixed, $L_T = \KL[q(\vx_T|\vx_0), p(\vx_T)]$ is a constant.
The terms $L_{t-1}$ for $t=2,...,T$ can be written as $L_{t-1} = \bE_{q(\vx_t|\vx_0)} [ \KL[q(\vx_{t-1}|\vx_t,\vx_0)\ | \ p(\vx_{t-1}|\vx_t)] ]$.
By further applying the reparameterization trick \citep{kingma2013auto}, the terms $L_{1:T-1}$ can be rewritten as a prediction of the noise $\epsilon$ added to $\vx_0$ in $q(\vx_t|\vx_0)$.
Parameterizing $\mu_{\theta}$ using the noise prediction $\epsilon_{\theta}$, we can write
\begin{align}
    \label{eq:Lt_simple}
    L_{t-1, \epsilon} &= \bE_{q(\epsilon)} \left[ w_t \|\epsilon_{\theta}(\sqrt{\bar{\alpha}_t}\vx_0 + \sqrt{1-\bar{\alpha}_t} ~ \epsilon, t) -\epsilon\|_2^2 \right] + \text{const.}
\end{align}
where $w_t = \frac{\beta_t^2}{2\sigma_t^2\alpha_t (1-\bar{\alpha}_t)}$,
which corresponds to the ELBO objective. The weights $w_t$ can also be written in terms of signal-noise-ratio as proposed in~\citep{kingma2021variational}. Empirically~\citep{ho2020} shows superior sample quality and stable training when using a re-weighted ELBO objective using $w_t=1$ with a predictable drop in likelihood performance. We call this loss $L_{\epsilon}$ in this paper.

\section{Few-Shot Diffusion Models}

\begin{figure}[ht]
\vspace*{-\baselineskip}
    \begin{subfigure}[b]{0.38\columnwidth}
        \centering
        \includegraphics[width=.8\linewidth,valign=t]{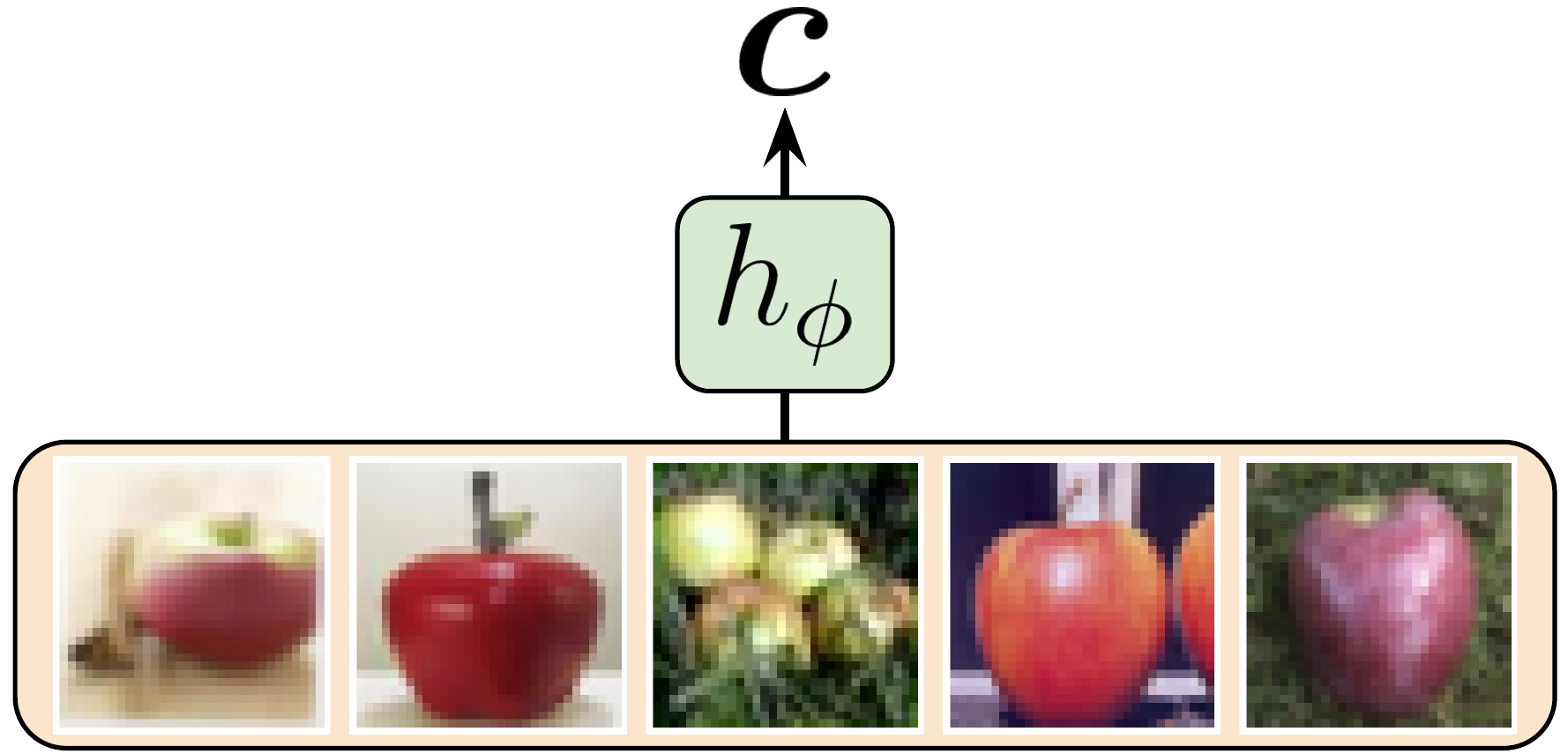}
        \caption{Context net $h_{\phi}$: Takes the set $\vX$ as input and outputs the context $\vc$.}
    \end{subfigure}
    \hspace{0.01\columnwidth}
    \begin{subfigure}[b]{0.58\columnwidth}
        \centering
        \includegraphics[width=0.8\linewidth,valign=t]{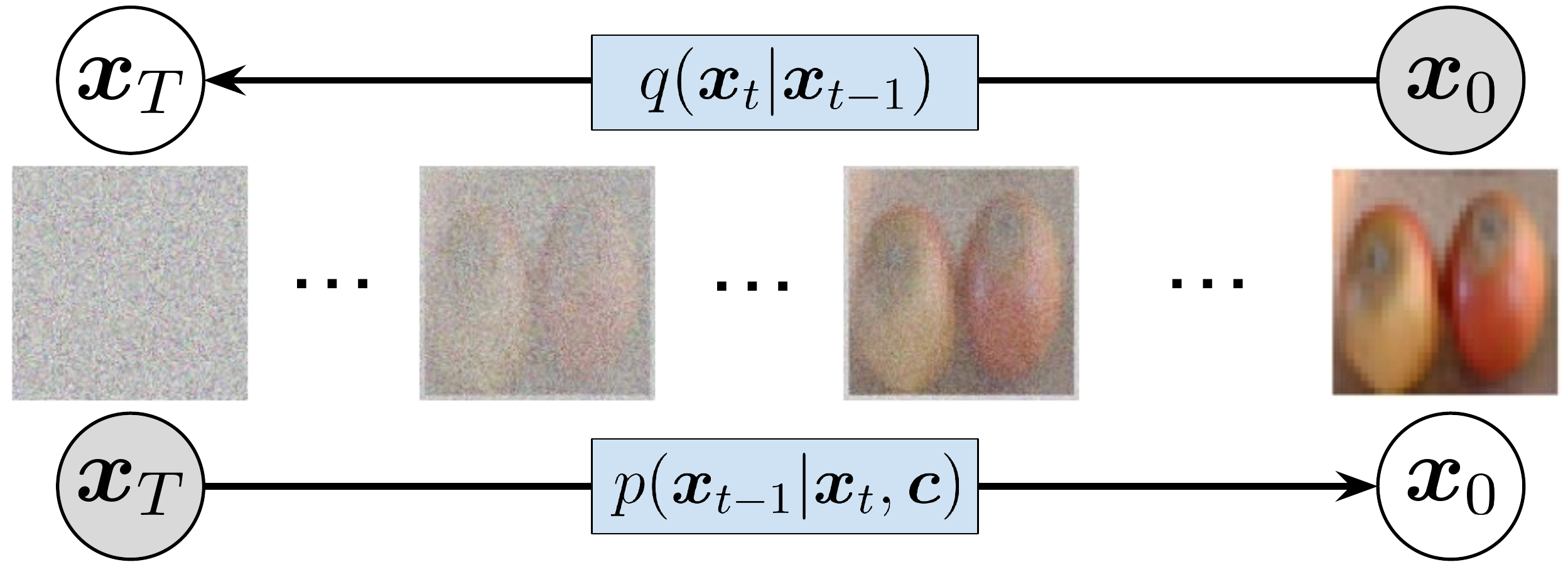}
        \caption{Conditional diffusion: A denoising diffusion model where the generative path is conditioned on the context $\vc$.}
    \end{subfigure}
    \vspace{4mm}
    \caption{Few-Shot Diffusion Models.}
    \label{fig:fsdm}
\end{figure}

In this paper, our goal is to learn to quickly adapt to new generation tasks. That is, we want to perform few-shot generation conditioned on a set $\vX$ containing previously unseen samples from a new task. We approach this problem using diffusion models: We learn a diffusion model $p_{\vtheta}(\vx | \vX)$ conditioned on the set $\vX$. We refer to our approach as Few-Shot Diffusion Models (FSDM). 
Our model can be broken down into two main parts: 1) A neural network $h_{\vphi}$ that produces a context representation $\vc = h_{\vphi}(\vX)$ of the set $\vX$, and 2) a conditional diffusion model that generates novel samples conditioned on the context $\vc$. See Fig. \ref{fig:fsdm} for an illustration.

\paragraph{Generative Model.} 
The generative model is a conditional diffusion model
\begin{equation}
    p_{\vtheta}(\vx_{0:T} | \vX) = p_{\vtheta}(\vx_T) \prod_{t=1}^T p_{\vtheta}(\vx_{t-1} | \vx_t, \vc), \quad \vc = h_{\vphi}(\vX),
\end{equation}
conditioned on the set $\vX$ through the context $\vc$ produced by the context net $h_{\vphi}$. In practice, we use a Vision Transformer (ViT~\citep{dosovitskiy2020image}) as the context net $h_{\vphi}$, but we also experiment with a UNet encoder. We discuss the use of ViT in Sec. \ref{sec:vit}. The generative path $p_{\vtheta}(\vx_{t-1} | \vx_t, \vc)$ is parameterized by a UNet, as is common practice for diffusion models. However, since we have an additional context $\vc$, we need the UNet to fuse the information in $\vx_t$ and $\vc$ to predict $\vx_{t-1}$. In this work we consider two main mechanisms for this, 1) a mechanism based on FiLM~\citep{perez2017film} and 2) we propose Learnable Attentive Conditioning (LAC), inspired by~\citep{rombach2021high}. We discuss these in greater detail in Sec. \ref{sec:cond}. The prior could also have been conditioned on $\vc$, i.e. $p_{\vtheta}(\vx_T|\vc)$, but for simplicity we use a standard unconditional Gaussian $p_{\vtheta}(\vx_T) = \cN(\vzero, \vI)$.

\paragraph{Inference Model.} 
Given the special structure of DDPM and FSDM, the inference model is parameter-free and we do not need to condition on $c$. This is a great simplification during training.
In practice FSDM employs a diffusion parameter-free posterior that degrades the information in the data at each step adding noise as presented in Eq.~\ref{eq:ddpm_inference_main}.

\paragraph{Loss and Training.} The negative ELBO can be expressed as a conditional version of Eq.~\ref{eq:loss_ddpm_main}, $L_{\mathtt{FSDM}} = L^c_0 + \sum_{t=2}^{T} L^{c}_{t-1} + L^{c}_T$. As for regular DDPMs, the loss can be decomposed into a sum of terms, one per layer, that can be computed independently. Training thus enjoys the same benefits where we can get efficient stochastic estimates of the objective by Monte Carlo sampling terms.
The conditional per-layer loss $L^{c}_{t-1}$ can then formulated as

\begin{equation}
    \label{eq:Lt_cond_simple}
    L^{c}_{t-1, \epsilon} = \bE_{q(\epsilon)} \left[ \|\epsilon_{\theta}(\vx_t, \vc) -\epsilon\|_2^2 \right], \quad \vx_t(\vx_0, \epsilon) = \sqrt{\bar{\alpha}_t}\vx_0 + \sqrt{1-\bar{\alpha}_t} ~ \epsilon.
\end{equation}

$L^{c}_T$ is unconditional and fixed in our model formulation, and $L^{c}_0$ is a negated conditional discretized normal likelihood, $L^{c}_0 = \bE_{q(\vx_1|\vx_0)} \left[- \log p_{\theta}(\vx_0 |\vx_1, \vc) \right]$. 

\begin{wrapfigure}[22]{r}{0.3\textwidth}
\vspace*{-\baselineskip}
    \centering
    \includegraphics[width=.25\textwidth]{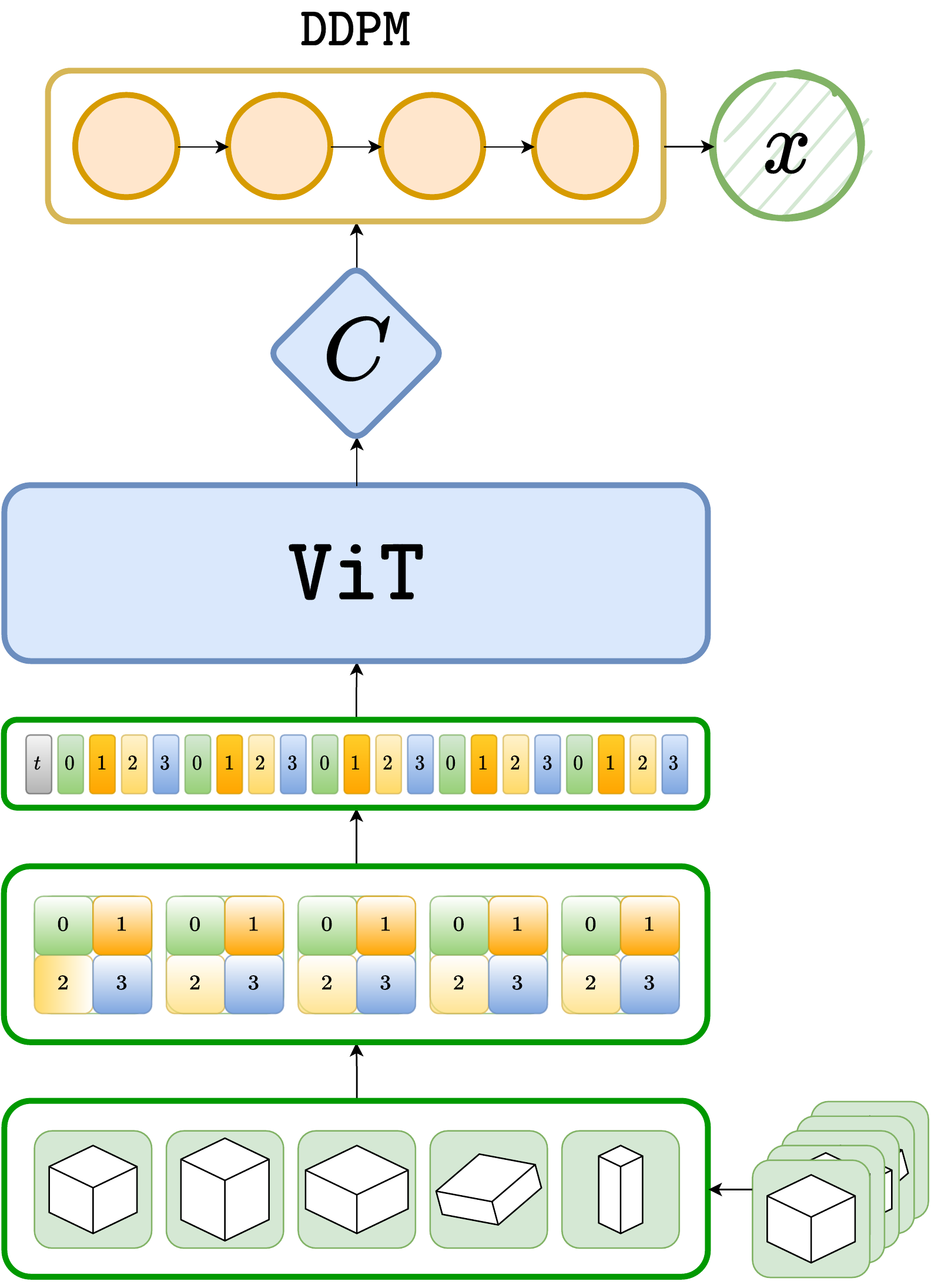}
    \caption{sViT architecture. 
    The input is a set $\vX$ of images. 
    These are split in non-overlapping patches and fed to a transformer encoder using a shared positional encoding, as indicated by the patch colors.
    The sViT outputs a context as a vector (V) or collection of visual tokens (T).
    The DDPM is conditioned on this information using FiLM or attention.}
    \label{fig:vis_vfsdm}
\vspace*{-\baselineskip}
\end{wrapfigure}
The context net extracts information from a small set $\vX$ which shares label with the input sample at hand $\vx$. 
In general during training the context can be input-dependent, $\vx \in \vX$ or input-independent, $\vx \notin \vX$: 
in the input-dependent scenario, given a set $\vX$, for each new input $\vx$ we can learn a different context aggregation. 
In the input-independent scenario, given the set $\vX$, the context aggregation is the same for any $\vx$.
We explore both context formulations during training. 
We noticed that the input-independent context works slightly better in-distribution but performs poorly in the out-distribution scenario, 
resulting in a conditioning mechanism that exploits wrong information at test time. 
The training with input-dependent context provides good performance out-distribution,
trading in-distribution sample variety for out-distribution conditioning quality. 
Following these findings we train the models with the input-dependent approach, where the aggregation mechanism works as a bottleneck.
At test time conditional and few-shot sampling is performed on a generic in or out-distribution set.

\subsection{ViT as Set Encoder} 
\label{sec:vit}
Transformers~\citep{vaswani2017attention} are the de-facto standard for natural language processing tasks and text generation~\citep{devlin2018bert, brown2020language}.
Recently Vision transformers (ViT,~\citep{dosovitskiy2020image}) unlocked the general power of attention for vision tasks.
However the use of transformers in latent variable models for generation is still limited. 
ViT gives us a flexible way to process images at the patch level. 
We adapt ViT to handle sets of images (sViT) similarly to~\citep{lee2021vision}: in particular we want to handle small sets (1-10) of images. 
The fundamental idea is that we want to extract global information from the set and each patch should contain global information for a specific region in the image (Fig.~\ref{fig:vis_vfsdm}).
In general we can condition the ViT encoder on the layer embedding $t$ using $\mathtt{ViT}(\vX, t ; \phi)$: doing so we obtain a cheap way to learn a per layer-dependent context, coarse for large $T$, and more refined for small $t$.
The use of tokens as input opens the door for a general domain-agnostic few-shot latent variable generator: our approach can be effortlessly employed for few-shot and conditional generation with any modality (text, speech, vision) simply tokenizing the input set and fine-tuning the patch embedding layer, without the need of any modification to the set-encoder, conditioning mechanism or generative process.

\subsection{Conditioning the Generative Process}
\label{sec:cond}
\begin{wraptable}{r}{0.4\textwidth}
\vspace*{-\baselineskip}
	\begin{center}
	\scriptsize
	\caption{Different conditioning mechanisms.
	}
		\begin{tabular}{lccc}
		\toprule
		 & $\vc$ & Cond \\
		\midrule
		FiLM & $\mathbb{R}^{d}$ & $m(\vc) \vu + b(\vc)$ \\
		LAC  & $\mathbb{R}^{{N} \times d}$ & $\mathtt{att}(\vu, \{\vc_p\}^{N_p}_{p=1})$\\
		\bottomrule
		\end{tabular}
		\label{tab:conditioning-mechanisms}
	\end{center}
\vspace*{-\baselineskip}
\end{wraptable}
After processing the patches and obtaining $\vc$, we need to find a way to condition the DDPM generative path.
We work with $\vc$ in two different forms: A \emph{vector} $\vc \in \mathbb{R}^d$ or a collection of $N$ \emph{tokens}, $\vc \in \mathbb{R}^{N \times d}$ as summarized in Table~\ref{tab:conditioning-mechanisms}.
In this work we consider two main conditioning mechanisms: FiLM~\citep{perez2017film} and Learnable Attentive Conditioning (LAC), inspired by~\citep{rombach2021high}. 

\paragraph{Vector (V).} 
One approach can be to condition the intermediate feature maps $\vu$ in the DDPM UNet  on $\vc$, for example using a FiLM like mechanism~\citep{perez2017film}, $\vu = m(\vc) \vu + b(\vc)$, where $m$ and $b$ are learnable and context-dependent.  
Given the special structure of the generative model, where all the layers share parameters $\theta$ and differ only through an embedding of the step $t$, the conditioning mechanism can be generically written as $\vu = f(\vu, \vc, t) = f(\vu_t, \vc_t)$. Merging together $\vc$ with the step embedding we can condition each layer, defining a generic per-step conditioning mechanism. In practice we found $\vu(\vc, t) = m(\vc) \vu(t) + b(\vc)$ being the best performing and flexible approach.

\paragraph{Tokens (T).} 
Alternatively, $\vc$ can be a collection of variables $\vc = \{\vc_{sp}\}^{N_s, N_p}_{s=1,p=1}$, where $N_p$ is the number of patches per sample, and $N_s$ the number of samples in the set. In this case, attention can be used to fuse information between the context $\vc$ and the feature maps $\vu$. In principle we could use the patches directly, i.e. $\vu = \mathtt{att}(\vu, \{\vc_{sp}\}^{N_s, N_p}_{s=1,p=1})$. 
However, this approach scales badly with the number of samples in the set $N_s$. 
Another option is to use a per-patch aggregation where we average of the set dimension $\vc_p = \tfrac{1}{N_s}\sum^{N_s}_{s=1} \vc_{sp}$ to obtain $N_p$ tokens $\{\vc_p\}^{N_p}_{p=1}$ that we feed to the ViT. 
We then use cross-attention~\citep{rombach2021high} on the per-patch averaged tokens 
$\vu = \mathtt{att}(\vu, \{\vc_p\}^{N_p}_{p=1})$ to condition DDPM. 
Using per-patch aggregation, we can process any number of samples without increasing the number of tokens used to condition DDPM and, more importantly, aggregate information from different samples in the context $\vc$.

\subsection{Variational FSDM} 
\label{sec:vfsdm}
Alternatively to our formulation of FSDM, we could specify a latent variable model where the context $\vc$ is a latent variable and the set $\vX$ is generated conditioned on $\vc$. We refer to this model as Variational FSDM (VFSDM) and write it like
\begin{equation}
    p_{\vtheta}(\vX_{0:T}, \vc) = p_{\vtheta}(\vc) \left[ \prod_{s=1}^S p_{\vtheta}(\vx_{0:T}^{(s)} | \vc) \right],\quad p_{\vtheta}(\vx_{0:T} | \vc) = p_{\vtheta}(\vx_T) \prod_{t=1}^T p_{\vtheta}(\vx_{t-1} | \vx_t, \vc).
\end{equation}
In this case, the inference model will be a combination of the parameter-free diffusion posterior and a parameterized encoder for $\vc$,
\begin{equation}
    q_{\phi}(\vX_{1:T}, \vc | \vX_0) = \underbrace{q_{\phi}(\vc | \vX_0)}_{\mathtt{Set\,Encoder}} \, \underbrace{ \left[ \prod_{s=1}^S q(\vx_{1:T}^{(s)} | \vx_0^{(s)}, \vc) \right] }_{\mathtt{Diffusion}}, \quad q(\vx_{1:T} | \vx_0, \vc) = \prod_{t=1}^T q(\vx_t | \vx_{t-1}, \vc).
\end{equation}
Furthermore, the negative ELBO will contain an extra KL term between the encoder $q_{\vphi}(\vc | \vX_0)$ and the prior $p(\vc)$,
\begin{equation}
    L_{\mathtt{VFSDM}} = \bE_{q_{\vphi}(\vc | \vX_0)} [L_{\mathtt{FSDM}}] +~\mathbb{KL}\left[q_{\vphi}(\vc | \vX_0) \| p_{\vtheta}(\vc)\right]
\end{equation}
We originally worked with this model, but found the training to be more challenging, resulting in under-performance or poor conditioning properties.

\section{Experiments}
\paragraph{Setup.} 
We use as backbone the standard DDPM model proposed in~\citep{ho2020, nichol2021} with fixed $T=1000$ and linear $\beta$ schedule.
We reduce the number of channels in the model to 64 obtaining a 25M parameter model.
We train using $L = L_{\epsilon} + \lambda L_{\mathtt{vlb}}$ with $\lambda=0.001$ from~\citep{nichol2021} that we found gave us a good balance between sample quality and training stability.
We employ a Unet (10M) and a ViT (5M) as set encoders.
We use the standard unconditional DDPM and conditional DDPM variants as baselines.
In general our approach can be applied to condition any unconditional diffusion model~\citep{ho2020, kingma2021variational, song2020score}. For this reason we limit our attention to standard DDPM models with a discrete number of layers. 

There is little work on few-shot generation with diffusion models. The field focused mostly on adapting unconditional diffusion models at sampling time~\citep{choi2021ilvr} instead of learning parametric conditional diffusion models. 
Most conditional models focus on a specific applications. Here we consider the general case of few-shot generation for unknown classes at test time. 

Previous work~\citep{sinha2021d2c} proposed few-shot conditional attribute generation, adapting an unconditional model and fine-tuning a parametric encoder on a specific new attribute (for example conditioning a diffusion model, trained on faces without glasses, on glasses): however where they use 100+ samples to condition a distribution on a specific attribute, our goal is to few-shot novel realistic, complex classes with 1-10 samples without relying on pre-trained models but explicitly conditioning the DDPM learning dynamics. 

\paragraph{Baselines.}
We compare \underline{\texttt{FSDM}} with unconditional and conditional baselines. 
For the conditional baselines, we adapt conditional diffusion models in the literature~\citep{dhariwal2021diffusion, sinha2021d2c, vahdat2021score} to the few-shot class generation scenario.
We use a \underline{\texttt{DDPM}}~\citep{ho2020, nichol2021} as unconditional baseline. 
We then compare with two main conditional diffusion models: a \underline{\texttt{cDDPM}}, where a Unet encoder~\citep{ronneberger2015u} processes independently the images in $\vX = \{x_s\}^S_{s=1}$ and aggregates using a mean operator, $\vc = \tfrac{1}{S} \sum^{S}_{s=1} f_{\phi}(x_s)$; the Unet encoder has the same structure of the guiding classifier in~\citep{dhariwal2021diffusion} used to learn class-conditional models.  
We then consider a \underline{\texttt{sDDPM}} adapting ideas in~\citep{sinha2021d2c} without contrastive learning, where a ViT encoder~\citep{dosovitskiy2020image} splits $\vX$ in patches and processes them jointly, as depicted in Fig.~\ref{fig:vis_vfsdm}, and aggregates all the patches using a mean operator, $\vc = \tfrac{1}{N_s N_p} \textstyle\sum^{N_s}_{s=1} \sum^{N_p}_{p=1} \vc_{sp}$.
We also train a variational variant inspired by~\citep{vahdat2021score}, \underline{\texttt{vDDPM}}, that uses standard amortized variational inference~\citep{jordan1999introduction, kingma2013auto} on the per-set latent variable.
For cDDPM, sDDPM and vDDPM, $\vc$ is a vector and the conditioning mechanism is FiLM based~\citep{perez2017film}.
We also compare explicit conditioning with test-time (or sampling-time) conditioning as proposed in~\citep{choi2021ilvr}.

We consider also two variants for FSDM, called \underline{\texttt{FSDM-s}} and \underline{\texttt{vFSDM}}: 
FSDM-s employees a different way to extract and aggregate set-information using ViT: we stack all the samples on the channel dimension and process them as one entity as proposed in~\citep{lee2021vision}. vFSDM is a variational formulation where we learn a distribution over a set of tokens in the conditioning mechanism and we deal with amortized variational inference with quantized latents. We adapt the relaxation proposed in~\citep{ramesh2021zero, nichol2021glide} and the vector quantization proposed in~\citep{oord2018representation, razavi2019generating} to learn this model.

\paragraph{Datasets.} 
We extensively test the baselines and our approach on 4 image datasets with different complexity and size: \underline{\texttt{Omniglot}} (28)~\citep{lake2011one} using the binarization provided in~\citep{bartunov2018few}. 
\underline{\texttt{FS-CIFAR100}} (32)~\citep{oreshkin2018tadam} using the original class split and mixing all the classes together (\texttt{CIFAR100mix}).
\underline{\texttt{miniImageNet}} (32)~\citep{vinyals2016matching, ravi2016optimization} dataset to test few-shot and transfer capacity. \underline{\texttt{CelebA}} (64)~\citep{liu2018large} for additional visualizations. We refer to FS-CIFAR100 as CIFAR100 in the following.

\subsection{Few-Shot Generation} 
We compare the generative models in terms of denoising capacity, summing over $T$ steps Eq.~\ref{eq:Lt_cond_simple}, FID~\citep{heusel2017gans} for sample quality, sFID~\citep{nash2021generating} to capture spatial relationships, Precision and Recall~\citep{kynkaanniemi2019improved} for measuring variety and mode coverage. 
We consider two main scenarios: in-distribution (In), testing on classes seen during training; and out-distribution (\underline{Out}), testing on classes unknown during training (the few-shot scenario). 
We perform qualitative experiments on Omniglot, CIFAR100 and CelebA, and quantitative experiments on CIFAR100 and miniImageNet in Table~\ref{tab:generative_metrics} and Table~\ref{tab:ilvr}.

In Table~\ref{tab:generative_metrics} FSDM outperforms the unconditional and conditional baselines on both datasets and scenarios, providing evidence that a token-based representation jointly with cross-attention conditioning are effective mechanisms for few-shot generation.
FSDM is a better denoiser ($L_{\epsilon}$) and image generator (FID, sFID, P, R) than strong conditional baselines.
We notice that in-distribution, DDPM and the conditional baselines (cDDPM and sDDPM) perform well as expected, but tend to under-perform out-distribution. This is expected for unconditional DDPM. 
For the conditional variants global aggregation is not expressive enough to represent complex novel realistic classes.
Processing the set using FSDM-s tends to work better than the baselines but under-performs FSDM.
Additionally, FSDM is an efficient learner, being able to extract more information from less data and converge faster than the baselines (Fig.~\ref{fig:FSDM_mse_cifar100mix}).
\begin{table}
\vspace*{-\baselineskip}
\begin{center}
\scriptsize
\caption{Few-Shot generative evaluation on different datasets and for different metrics.
We test few-shot generation on CIFAR100 and miniImageNet and transfer from CIFAR100 to MinimageNet.
In CIFAR100 we use the original split and all the test classes are from new categories.
In: in-distribution - we evaluate the models on known classes. 
The context $\vc$ can be:
V: deterministic-vector. 
T: deterministic-tokens. 
vV: variational-vector.
vT:  variational-tokens.
Out: out-distribution - we evaluate the models on unknown classes (few-shot task).
$L_{\epsilon}$: denoising loss; FID: Frechet score; sFID: spatial FID; P: precision; R: recall.
We do not use augmentation to train these models. 
We use 10K samples for the metrics and 250 steps.
LAC: learnable attentive conditioning.\\
FSDM performs better than the baselines on known and unknown classes, providing evidence that the token-based representation and the cross-attention conditioning mechanism are effective for few-shot generation in diffusion models.
}
    \setlength\tabcolsep{6.0pt}
    \begin{tabular}{lccc cccccccccc}
    \toprule
    & &&
    & \multicolumn{2}{c}{ $\downarrow$ $L_{\epsilon}$}
    & \multicolumn{2}{c}{$\downarrow$ FID} & \multicolumn{2}{c}{$\downarrow$ sFID} & \multicolumn{2}{c}{$\uparrow$ P} & \multicolumn{2}{c}{$\uparrow$ R}\\
    & Enc & Cond & $\vc$ & In & \underline{Out} & In & \underline{Out} & In & \underline{Out} & In & \underline{Out} & In & \underline{Out} \\
    \\
    \emph{Few-Shot Generation} \\
    \textbf{CIFAR100} (32) \\
    \midrule
    \vspace{4pt}
    DDPM & -      & -     &  -  & 6.92 & 8.14 & 15.35 & 62.84    & 18.03 & 28.91 & 0.66 & 0.58 & 0.56 & 0.40   \\
    cDDPM   & Unet   & FiLM  &  V  & 6.58 & 8.08 & 11.84 & 38.50& 17.64 & 22.21 & 0.70& 0.55& 0.56 & 0.46 \\
    vDDPM         & Unet   & FiLM  &  vV & 7.15 & 8.03 & 14.98 & 62.58& 17.80 & 27.50 & 0.65 & 0.58 & 0.56 & 0.41  \\
    \vspace{4pt}
    sDDPM                                   & ViT    & FiLM  &  T  & 6.70 & 8.17 & 13.34 & 45.50 & 21.32 & 29.87 & 0.67 & 0.54& 0.55 & 0.46 \\
    vFSDM (Ours)                                                 & ViT    & LAC   &  vT & 6.94 & 8.00 & 13.63 & 63.73 & 17.55 & 28.85 & 0.65 & 0.58 & 0.58 & 0.38 \\
    FSDM-s (Ours)                                                & ViT    & LAC &  T  & 5.81 & 7.72 & 12.39 & 40.71 & \textbf{17.26} & 22.12 & 0.68 & 0.57 & 0.57 & 0.44 \\
    \textbf{FSDM (Ours)}                                         & ViT    & LAC   &  T  & \textbf{5.56} & \textbf{6.88} & \textbf{10.21} & \textbf{35.07} & \textbf{17.48} & \textbf{20.95} & \textbf{0.72} & \textbf{0.62}& \textbf{0.65} & \textbf{0.53}  \\
    \\
    \emph{Few-Shot Generation} \\
    \textbf{miniImageNet} (32) \\
    \midrule
    \vspace{4pt}
    DDPM & -      & -     &  -  & 9.73 & 10.08 & 22.84& 41.37 & \textbf{20.01} & 23.37 & 0.60 & 0.58 & 0.54 & 0.47  \\
    cDDPM   & Unet   & FiLM  &  V  & 9.50 & 10.12 & 17.47 & 32.22 & 20.04 & \textbf{21.57} & 0.65 & 0.59 & 0.55 & 0.52\\
    \vspace{4pt}
    sDDPM           & ViT    & FiLM  &  V  & 9.44 & 10.18& 18.21 & 35.86& 20.92 & 22.49& 0.64 & 0.56& 0.53 & 0.49\\
    FSDM-s (Ours)                                                & ViT    & LAC &  T     & 8.46 & 9.47 & 22.40 & 35.83 & 20.79 & 22.19 & 0.65 & 0.59 & 0.52 & 0.52 \\
    \textbf{FSDM (Ours)}                                         & ViT    & LAC   &  T   & \textbf{7.76} & \textbf{8.30} & \textbf{15.39} & \textbf{30.62} & \textbf{19.83} & \textbf{21.84} & \textbf{0.67} & \textbf{0.64} & \textbf{0.61} & \textbf{0.56} \\
    \\
    \emph{Transfer} \\
    \textbf{CIFAR100} (32) \\
    $\quad\quad\downarrow$ \\
    \textbf{miniImageNet} (32) \\
    \midrule
    \vspace{4pt}
    DDPM & -      & -     &  -   &-& 10.68 &-& 63.13 &-& 33.23 &-& 0.61 &-& 0.30 \\
    cDDPM     & Unet   & FiLM  &  V   &-& 10.77 &-& 41.00 &-&  \textbf{25.61} &-& 0.59 &-& 0.39 \\
    \vspace{4pt}
    sDDPM                                   & ViT    & FiLM  &  V   &-& 10.85 &-& 47.73 &-& 32.90 &-& 0.56  &-& 0.37 \\
    FSDM-s (Ours)                                                 & ViT    & LAC   &  T   &-& 10.37 &-& 42.32 &-& \textbf{25.74} &-& 0.61 &-& 0.37 \\
    \textbf{FSDM (Ours)}                                          & ViT    & LAC   &  T   &-& \textbf{9.60} &-& \textbf{39.55}&-& 27.99&-& \textbf{0.65}&-& \textbf{0.45} \\
    \bottomrule
    \end{tabular}
\label{tab:generative_metrics}
\end{center}
\vspace*{-\baselineskip}
\end{table}

\subsection{Transfer}
The goal of FSDM is to perform few-shot generation on objects never seen during training. However there are multiple challenges when dealing with new classes, in particular if these new classes are from new categories and datasets.
Imagine to train a model on \texttt{cats} and \texttt{lions}, and then provide \texttt{tiger} at test time: even if the model has never seen a \texttt{tiger}, the encoder can extract information from a small set of tigers leveraging classes with similar animals.
In a way the model can "interpolate" between the set at hand and similar classes. 
But if we train on \texttt{apples} and \texttt{oranges} and test on \texttt{tiger}, the model is challenged in a more fundamental way. 
There is no way to interpolate with known classes and the model has to rely mostly on the conditioning set. 
Between these two extremes there is a spectrum of challenges in few-shot generation and we want to explore how far our model can adapt to new information. For this reason we test few-shot transfer on a different dataset. 
We take models trained on CIFAR100 and test them (without gradient based adaptation) on MinImageNet. This is a difficult generalization task. 
We report results in the bottom part of Table~\ref{tab:generative_metrics}. All the models struggle compared to new classes from the same dataset. However FSDM is still able to extract more information than the baselines, providing additional evidence that our framework can be used in a variety of small and large adaptation tasks.

\subsection{Sampling}
\begin{figure}
\vspace*{-\baselineskip}
    \centering
    \includegraphics[trim={0, 10cm, 0, 0},clip, width=.45\textwidth]{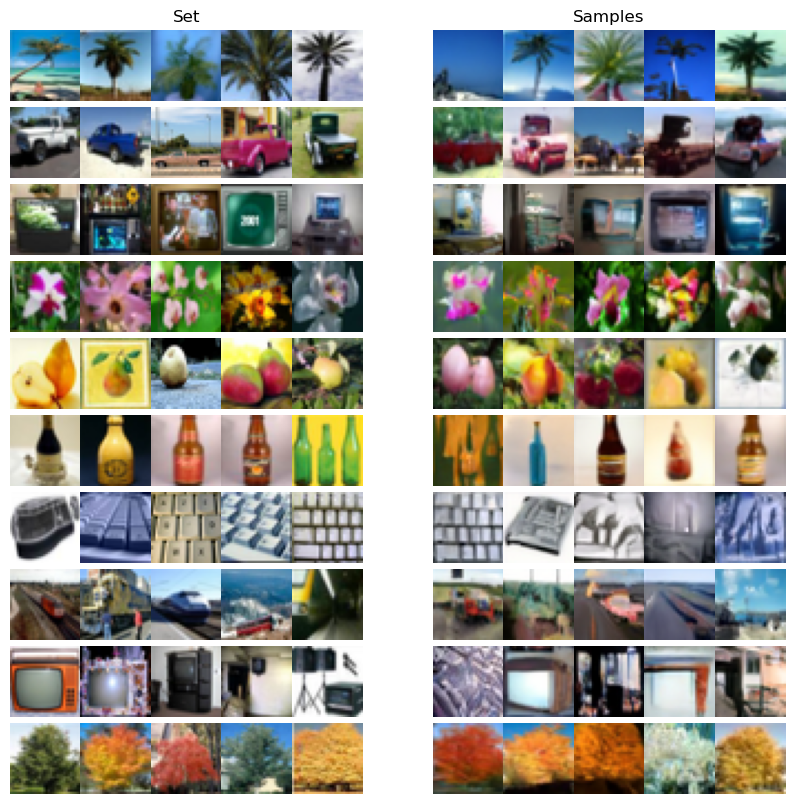}
    \quad\quad
    \includegraphics[trim={0, 10cm, 0, 0},clip, width=.45\textwidth]{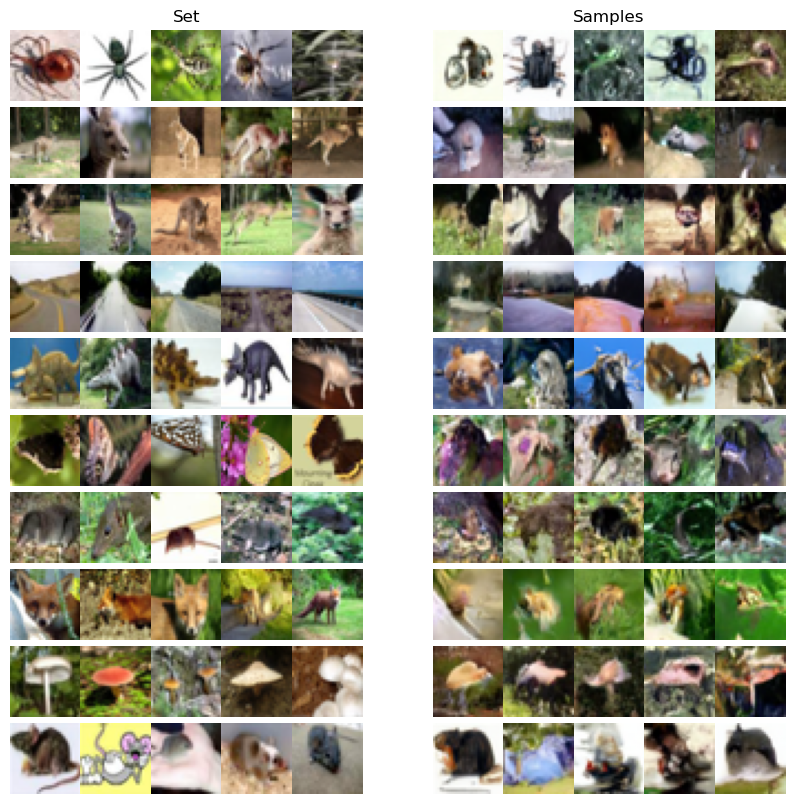}
    \caption{Few-Shot Conditional samples on CIFAR100 using a FSDM. Left side conditioning set and samples from in-distribution classes; right side conditioning set and samples from out-distribution classes.
    More samples in higher resolution in Appendix Fig.~\ref{fig:conditional_samples_cifar100_app} and Fig.~\ref{fig:few_shot_samples_cifar100_app}. 
    }
    \label{fig:conditional_samples_cifar100}
\end{figure}

\begin{wraptable}{r}{0.3\textwidth}
\scriptsize
\vspace*{-\baselineskip}
\setlength\tabcolsep{4.0pt}
	\begin{center}
	\caption{Few-Shot metrics test set (new classes) for different datasets. 
	We compare unconditional DDPM, test-time adaptation with ILVR and FSDM.}
		\begin{tabular}{lccc}
		\toprule
		       & $\downarrow$ FID & $\uparrow$ P & $\uparrow$ R  \\
		\textbf{CIFAR100} \\
		\midrule
		DDPM~\citep{ho2020}  & 62.84 & 0.58 & 0.40 \\
		ILVR~\citep{choi2021ilvr} & 45.83 & \textbf{0.62}  & 0.38 \\
		\textbf{FSDM (Ours)} & \textbf{35.07} & \textbf{0.62} & \textbf{0.53} \\
		\\
		\textbf{miniImageNet} \\
		\midrule
		DDPM~\citep{ho2020}  & 41.37 & 0.58 & 0.47 \\
		ILVR~\citep{choi2021ilvr} & 41.68 & 0.59  & 0.46 \\
		\textbf{FSDM (Ours)} & \textbf{30.62} & \textbf{0.64} & \textbf{0.56} \\
		\\
		\textbf{CIFAR100}      \\
        $\quad\quad\downarrow$ \\
        \textbf{miniImageNet}   \\
        \midrule
	    DDPM~\citep{ho2020}  & 63.13 & 0.61 & 0.30\\
		ILVR~\citep{choi2021ilvr} & 53.12 & 0.58 & 0.31 \\
		\textbf{FSDM (Ours)} & \textbf{39.55} & \textbf{0.65} & \textbf{0.45} \\
		\bottomrule
		\end{tabular}
		\label{tab:ilvr}
	\end{center}
\vspace*{-\baselineskip}
\end{wraptable}
FSGM can sample known and unknown classes conditioning on few samples.
When sampling known classes we are simply sampling conditional iid from a set $\vX$ summarized by a certain $\vc$.
When performing real few-shot sampling we condition on unknown classes using small sets of samples.
In Fig~\ref{fig:conditional_samples_cifar100} we show samples from known-classes (left panel) and from unknown-classes (right panel). The visual quality of the unknown classes is obviously worse than the known one. However the model can extract content information from few-samples and complex realistic classes in an effective way. We report additional visualizations on Omniglot and CelebA in Fig~\ref{fig:conditional_samples_celeba}.

\subsection{Test-time conditioning}
In this paper we argue the case that explicit adaptation during training is a powerful way to condition diffusion models. 
However inference-time adaptation~\citep{choi2021ilvr} has also been shown to be an effective mechanism to adapt diffusion models to new distributions and does not require retraining the model or a parametric encoder.
In Table~\ref{tab:ilvr} we compare the ILVR method, a powerful conditioning mechanism at sampling time, with FSDM, that condition the model during training, in the few-shot generation scenario, i.e. conditioning on samples from unknown classes during testing. 
We use the same datasets and evaluation procedure proposed in Table~\ref{tab:generative_metrics}.
We see that ILVR improves the result compared to a standard unconditional DDPM in terms of few-shot generation. However the task is challenging and the adaptation we require is on new classes and not only on new attributes. FSDM outperforms ILVR in this setting, providing additional evidence that explicit conditioning during training is essential for few-shot generation of realistic and complex objects.
\section{Related Work}

\paragraph{Few-Shot Latent Variable Models.}
Historically the machine learning community has focused its attention on supervised few-shot learning~\citep{lake2011one, lake2015human}, solving a classification or regression task on new classes at test time given a small number of labeled examples.
The problem can be tackled using metric based approaches~\citep{vinyals2016matching, snell2017prototypical, oreshkin2018tadam, sung2018learning, shyam2017attentive}, 
gradient-based adaptation~\citep{finn2017model, andrychowicz2016learning, hochreiter2001learning}, optimization~\citep{ravi2016optimization}, and posterior inference~\citep{garnelo2018conditional, gordon2019convolutional, requeima2019fast, grant2018recasting, ravi2018amortized}.
More generally, the few-shot learning task can be recast as Bayesian inference in hierarchical modelling~\citep{grant2018recasting, ravi2018amortized}. 
In such models, typically parameters or representations are conditioned on the task, and conditional predictors are learned for such task.
In~\citep{xu2019metafun} an iterative attention mechanism is used to learn a query-dependent task representation for supervised few-shot learning.
Modern few-shot generation in machine learning was introduced in~\citep{lake2011one}. 
The Neural Statistician~\citep{edwards2016towards} is one of the first few-shot learning models in the context of VAEs~\citep{kingma2013auto, rezende2014stochastic}.
The model has been improved further increasing expressivity for the conditional prior using powerful autoregressive models~\citep{hewitt2018variational}, a non-parametric formulation for the context~\citep{wu2020meta}, hierarchical learnable aggregation for the input set~\citep{giannone2021hierarchical}, and exploiting supervision~\citep{garnelo2018neural}.
~\citep{rezende2016one} proposed a recurrent and attentive sequential generative model for one-shot generation based on~\citep{gregor2015draw}.
Powerful autoregressive decoders and gradient-based adaptation are employed in~\citep{reed2017few} for one-shot generation. 
The context $\mathbf{c}$ in this model is a deterministic variable.
In GMN~\citep{bartunov2018few} a variational recurrent model learns a per-sample context-aware latent variable. 

\paragraph{Vision Transformers.}
Transformers~\citep{vaswani2017attention} have shown remarkable performance on unstructured text based data.
Vision Transformers (\citep{dosovitskiy2020image}, ViT) have recently emerged as a transformer variant to process image-like data using a patch-based approach. Then these patches are encoded as tokens and fed to a standard transformer encoder.
ViT has been used for discriminative tasks with remarkable results~\citep{touvron2021training, zhou2021deepvit, caron2021emerging, liu2021swin}. 
However less work has been done to use ViT in the context of generative latent variable models.
Recently masked autoencoders~\citep{he2021masked}, based on the ViT formulation, have been proposed for self-supervised learning and pretraining~\citep{bao2021beit}.
ViT variants for small and little data~\citep{lee2021vision, liu2021efficient, hassani2021escaping} have been proposed and our patch aggregation relies on similar ideas.  

\paragraph{Conditional Diffusion Models.}
The standard DDPM~\citep{ho2020} can be improved using likelihood-based training~\citep{durkan2021maximum, huang2021variational}, continuous time modeling~\citep{song2020score, vahdat2021score}, learnable noise scales~\citep{kingma2021variational}, efficient sampling mechanism~\citep{song2020denoising, bao2022analytic, jing2022subspace, salimans2022progressive, jolicoeur2021gotta, watson2021learning, kong2021fast}, and exploiting powerful (variational) autoencoders for dimensionality reduction~\citep{rombach2021high, preechakul2021diffusion, pandey2022diffusevae}.
Methods to condition DDPM have been proposed, conditioning at sampling time~\citep{choi2021ilvr}, learning a class-conditional score~\citep{song2020score}, explicitly conditioning on class information~\citep{nichol2021, nichol2021improved}, physical properties~\citep{xu2022geodiff, xie2021crystal, hoogeboom2022equivariant}, side information~\citep{baranchuk2021label, ho2022cascaded}, and temporal structure~\citep{ho2022video, harvey2022flexible}.
We present a more general class of methods to condition diffusion models based on set-conditioning: We learn a parametric conditioning mechanism at the set-level and a conditional diffusion process at the sample-level.
A conditional DDPM has been proposed for point-cloud generation~\citep{luo2021diffusion}. However they limit their attention to a specific application, where we consider the general idea to encode sets of generic data and use for few-shot generation and transfer.
Retrieval-based approaches~\citep{blattmann2022retrieval, ashual2022knn} use an external database and pre-trained contrastive embeddings~\citep{radford2021learning}. We similar increase expressivity using a collection of sample similar to the input selecting a set from the same class without relying on a retrieval mechanism, pre-trained model and an external database.
In~\citep{rombach2021high} DDPM leverages a large vector quantized~\citep{oord2018representation} pretrained autoencoder to encode the data in latent space, and such encoder can be used to condition on generic data. 
This approach is effective but expensive and rely on large pre-trained models on relevant datasets not always easily available.
VAE-DDPM models~\citep{vahdat2021score, pandey2022diffusevae} have been proposed to learn conditional models in latent space. Text-to-image diffusion models~\citep{nichol2021glide, ramesh2022hierarchical} have been recently proposed for guided generation. Our approach can be easily adapted to work with text tokens instead of visual tokens, simply changing the patch encoder.

\section{Conclusion}
We presented Few-Shot Diffusion Models, a flexible framework to adapt quickly to different generative processes at test-time, leveraging advances in Vision Transformers and Diffusion Models. We show how conditioning a diffusion model with rich, expressive information gives superior performance in a wide range of experiments in and out-distribution. Few-shot generation is performed on realistic, complex sets of images, showcasing a promising direction for large-scale few-shot latent variable generative models.

\section*{Acknowledgement}
We would like to thanks
Anders Christensen, Andrea Dittadi and
Nikolaos Nakis
for insightful comments and useful discussions.

\newpage
\small
\bibliographystyle{plain}
\bibliography{main}

\clearpage
\normalsize
\appendix

\section{Additional Experiments}

In this section we discuss additional experiments and visualizations for FSDM.

\begin{figure}[ht]
    \centering
    \includegraphics[width=.9\linewidth]{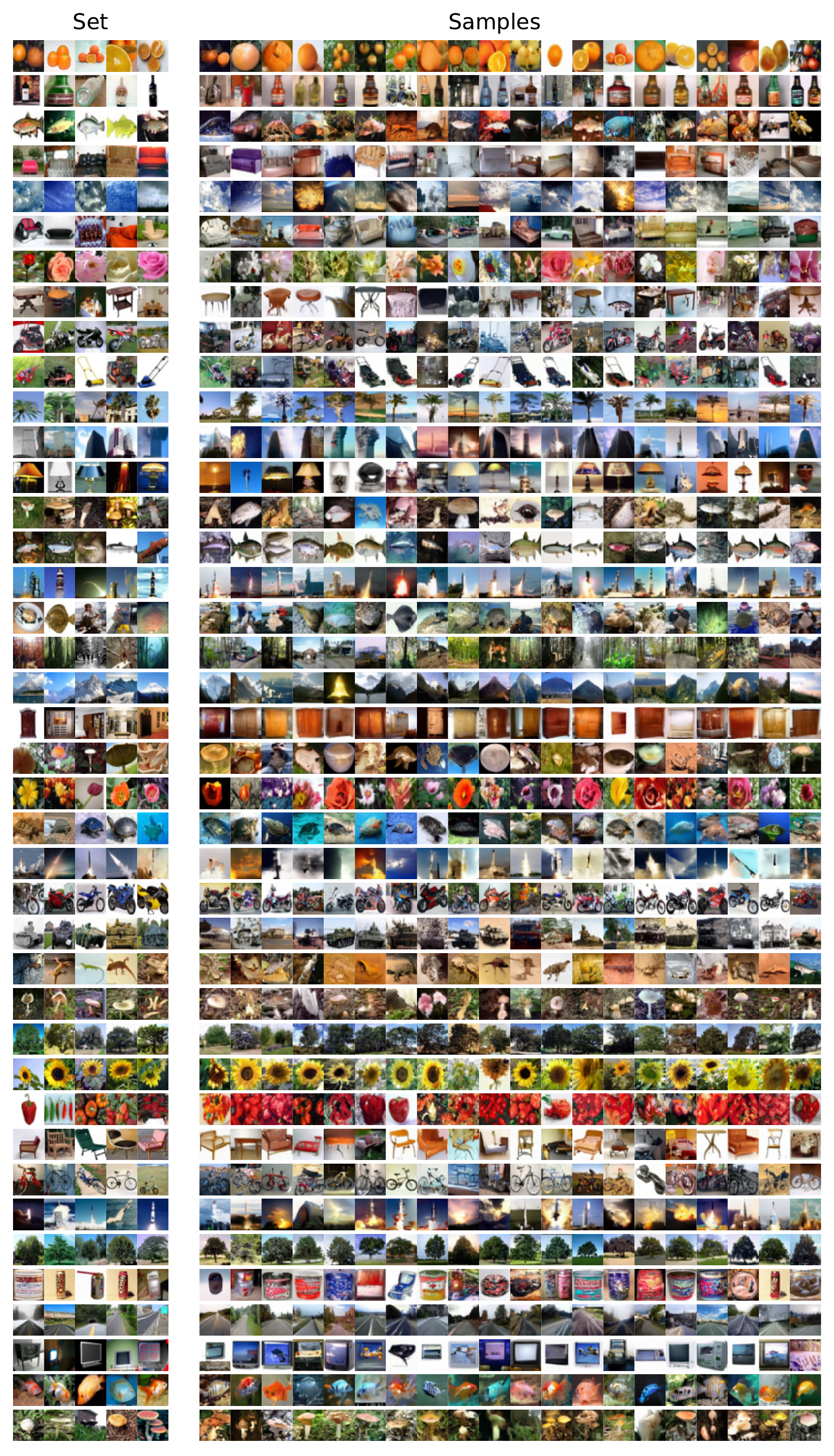}
    \caption{Set (left) and conditional samples (right) on CIFAR100 using a Few-Shot Diffusion Models.
    FSDM can extract content information from an handful of realistic examples and generate rich and complex samples from a variety of conditional distributions.}
    \label{fig:intro_samples_cifar100_app}
\end{figure}

Fig.~\ref{fig:intro_samples_cifar100_app} shows conditioning sets with cardinality 5 (left) and 20 conditional samples (right) for a FSDM using a large number of in-distribution classes from CIFAR100. We can see that samples are high quality and have large variability.

\begin{figure}[ht]
    \centering
    \includegraphics[width=.7\textwidth]{img/vis_model150000_indistro.png}
    
    \includegraphics[width=.7\textwidth]{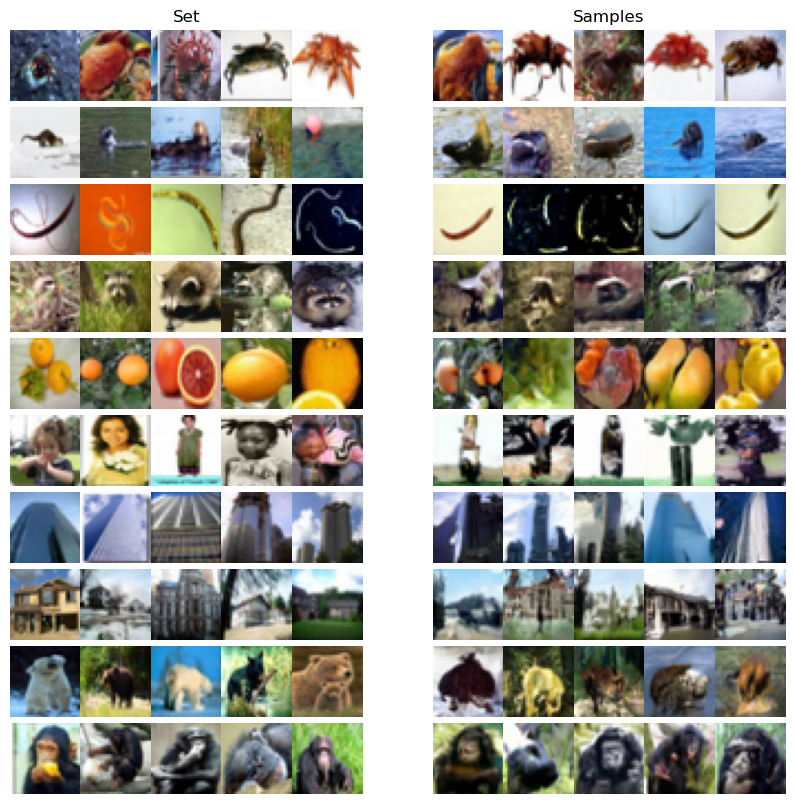}
    
    \caption{Conditional samples on CIFAR100 and CIFAR100mix using a FSDM. Left side conditioning set. Right side samples.
    Top in-distribution (known classes) on CIFAR100. 
    Bottom in-distribution (known classes) on CIFAR100mix.}
    \label{fig:conditional_samples_cifar100_app}
\end{figure}

\begin{figure}[ht]
    \centering
    \includegraphics[width=.7\textwidth]{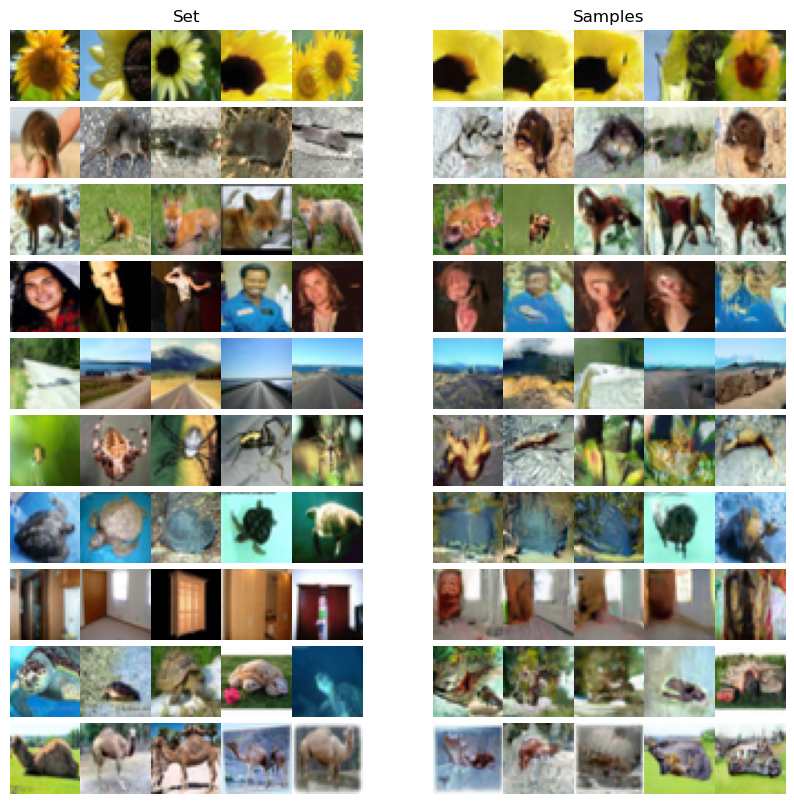}
    
    \includegraphics[width=.7\textwidth]{img/vis_cifar100mix_model100000_outdistro.png}
    
    \caption{Few-Shot samples on CIFAR100 and CIFAR100mix using a FSDM. Left side conditioning set. Right side samples.
    Top out-distribution for CIFAR100 (unknown classes from \underline{unknown} category). 
    Bottom: out-distribution for CIFAR100mix (unknown classes from \underline{known} category) .}
    \label{fig:few_shot_samples_cifar100_app}
\end{figure}

In Fig.~\ref{fig:conditional_samples_cifar100_app} we compare conditional samples on in-distribution classes using CIFAR100 (top) and CIFAR100-mix (bottom).
In Fig.~\ref{fig:few_shot_samples_cifar100_app} we perform the same experiment on out-of-distribution classes.
When using CIFAR100 the out-of-distribution sets are not only from novel classes never seen during training but also from new categories, i.e. training on \texttt{cats} and testing on \texttt{cars}.
When using CIFAR100-mix all the classes are mixed and the out-of-distribution sets are from new classes but not necessarily from novel categories, i.e. training on \texttt{cats} and testing on \texttt{tigers}.

Sample quality and variety decrease for out-of-distribution samples as expected. 
When FSDM is presented with out-of-distribution sets from CIFAR100, the model cannot rely on similar classes and few-shot samples have reduced variability. But when presented with out-of-distribution sets from CIFAR100mix, the model can rely on similar classes and the few-shot generation task is easier, giving rise to better samples.
Smarter conditioning mechanisms and expressive set-level latent variables can help in improving generalization and we will explore such possibilities in future work.

\begin{figure}[ht]
    \centering
    \includegraphics[width=\linewidth,valign=t]{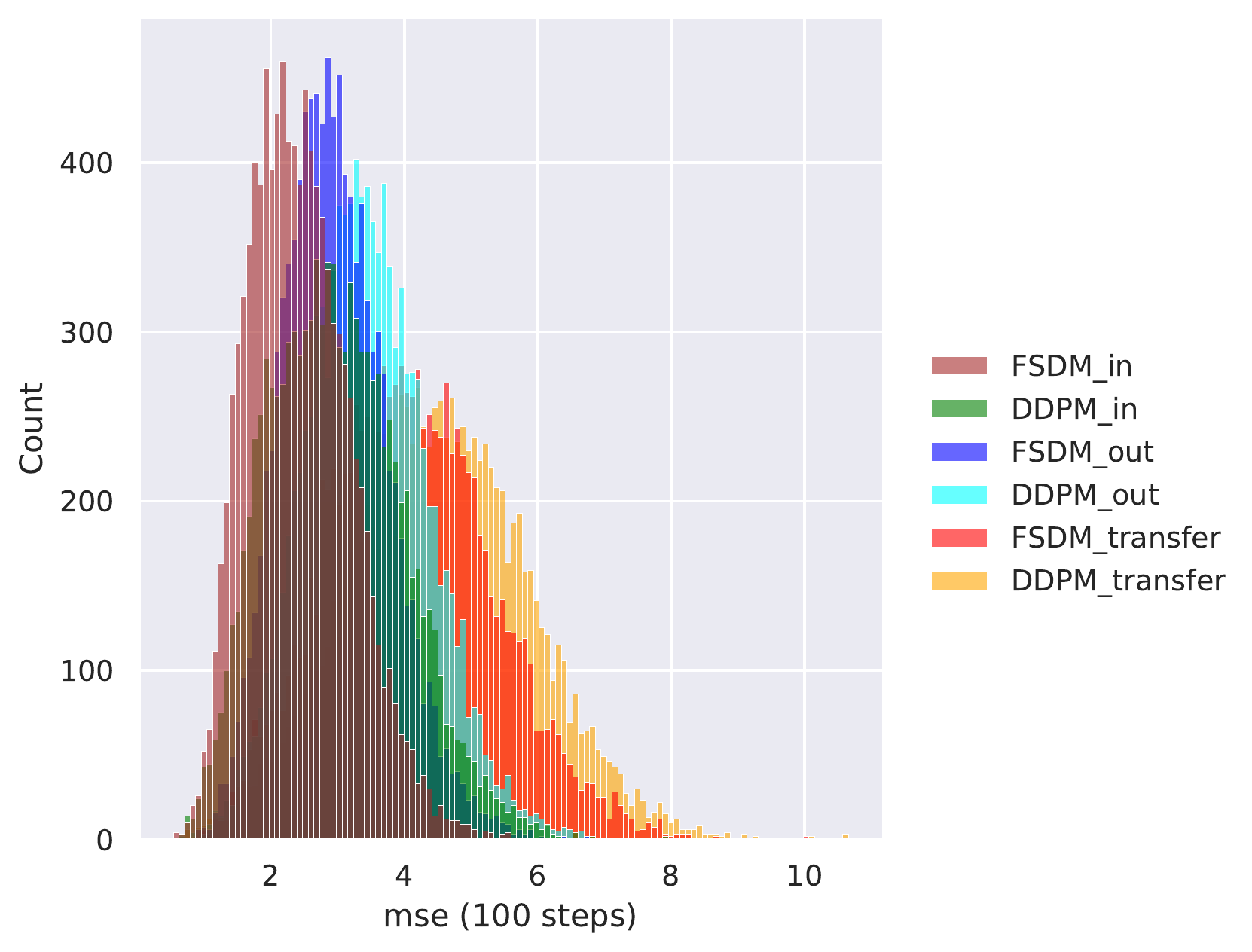}
        \caption{CIFAR100 in-distro, out-distro, transfer using $L_{\epsilon}$ computed with 100 steps of denoising.}
    \label{fig:bpd-hist}
\end{figure}

In Fig.~\ref{fig:bpd-hist} we use $L_{\epsilon}$ as a proxy signal to evaluate the capacity of DDPM and FSDM to distinguish between in-distribution and out-of-distribution samples. 
FSDM performs better than DDPM for this task, consistently distinguishing better between in-distribution and out-of-distribution classes.

\begin{figure}[ht]
    \centering
    \includegraphics[width=.7\textwidth]{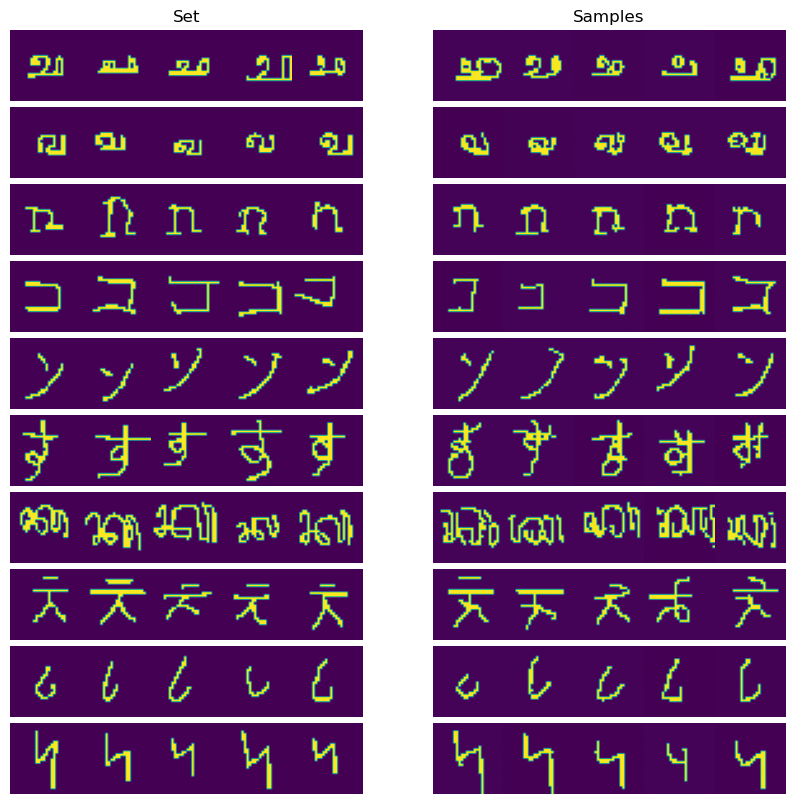}
    
    \includegraphics[width=.7 \textwidth]{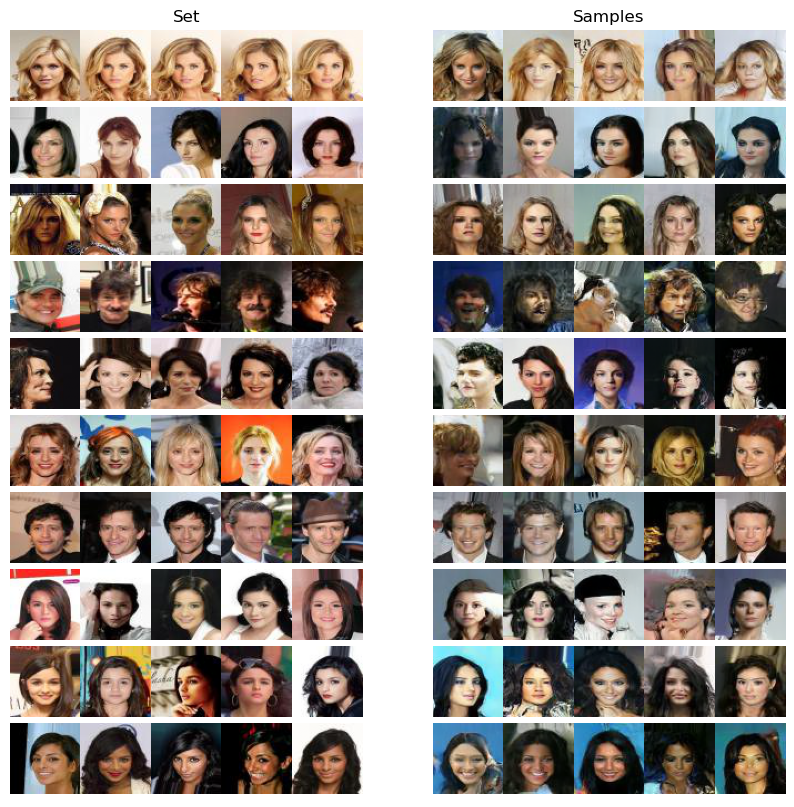}
    \caption{Conditional samples on Omniglot and CelebA using a FSDM with deterministic context. Random samples. We train the models for only 100K iterations with batch size 32. Even training for such short time, we can see that FSDM can extract content information from small complex sets and condition the generative path in a consistent way.
    Left side conditioning set; right side samples.}
    \label{fig:conditional_samples_celeba}
\end{figure}

In Fig.~\ref{fig:conditional_samples_celeba} we show visualizations for Omniglot and CelebA using short training runs. We train FSDM only for 100K iterations. The goal is to explore how fast the conditioning mechanism can extract and aggregate context information. As expected for a simple dataset as Omniglot the conditioning quality is high. 
For CelebA the conditioning mechanism struggles using so few training iterations. However, most of the context information is extracted successfully and samples are compatible with the content in the conditioning sets.

\clearpage
\section{Experimental Details}

\begin{table}[ht]
\begin{center}
\caption{Relevant Hyperparameters for FSDM. $L_{\texttt{h}}$: Loss hybrid~\cite{nichol2021improved}.}
\begin{tabular}{lcccc} 
\toprule
& Omniglot & CelebA & FS-CIFAR100 & miniImageNet\\ 
\midrule
Dimension               &  1x28x28   &  3x64x64       &  3x32x32 &  3x32x32\\
Number classes          &  1623      &  6349          & 100      & 96     \\
Classes Train           &  1000      &  4444          & 60       & 60     \\
Classes Val             &  200       &  635           & 20       & 16     \\
Classes Test            &  423       &  1270          & 20       & 20     \\
\midrule
Batch size              &  32        &  16            & 32       & 32     \\
Channels $\vc$          &  128       &  128           & 128      & 128    \\
Channels model          &  64        &  64$\div$128  & 64      & 64    \\
Channel multiplier      & (1, 1, 2, 4) & (1, 1, 2, 4) & (1, 1, 2, 4) & (1, 1, 2, 4) \\
Channels $\vz$          &  64        &  64            & 64      & 64           \\
Classes per set         &  1         &  1             &  1     & 1        \\
Heads                   &  12        &   12           & 12   & 12         \\
Iterations              &  100K      &  100K          & 200K & 200K       \\ 
Layers                  &  6         &   6            & 6    & 6          \\
Learning rate           &  $2e^{-4}$        &  $2e^{-4}$        & $2e^{-4}$ & $2e^{-4}$ \\
Likelihood              &  $\mathcal{N}$    &  $\mathcal{N}$    & $\mathcal{N}$     & $\mathcal{N}$  \\
Loss                    &  $L_{\texttt{h}}$ &  $L_{\texttt{h}}$ & $L_{\texttt{h}}$ & $L_{\texttt{h}}$ \\
Optimizer               &  Adam      &  Adam          &  Adam  &  Adam              \\
Set size          &  5         &   5            &  5   & 5          \\
\bottomrule
\end{tabular}
\end{center}
\end{table}

\end{document}